\documentclass[nohyperref]{article}

\usepackage{microtype}
\usepackage{graphicx}
\usepackage{subfigure}
\usepackage{booktabs} 

\usepackage{hyperref}

\graphicspath{{figures/}}

\usepackage[accepted]{icml2023_preprint}

\usepackage{amsmath}
\usepackage{amssymb}
\usepackage{mathtools}
\usepackage{amsthm}

\usepackage{pifont}
\usepackage{nicefrac}

\usepackage{localmacros} 

\usepackage[capitalize,noabbrev]{cleveref}

\theoremstyle{plain}
\newtheorem{theorem}{Theorem}[section]

\theoremstyle{definition}
\newtheorem{lemma}[theorem]{Lemma}
\newtheorem{definition}{Definition}[section]
\newtheorem{proposition}{Proposition}

\newtheorem{example}{Example}[section]
\newtheorem{environment}{Environment}
\newtheorem{remark}{Remark}[section]
\theoremstyle{remark}

\numberwithin{equation}{section}

\usepackage[disable,textsize=tiny]{todonotes}

\icmltitlerunning{FLEX: an Adaptive Exploration Algorithm for
Nonlinear Systems}

\begin{document}

\twocolumn[
    \icmltitle{FLEX: an Adaptive Exploration Algorithm for
        Nonlinear Systems}



    \icmlsetsymbol{equal}{*}

    \begin{icmlauthorlist}
        \icmlauthor{Matthieu Blanke}{argo}
        \icmlauthor{Marc Lelarge}{argo}
    \end{icmlauthorlist}

    \icmlaffiliation{argo}{INRIA, DI ENS, PSL Research University}

    \icmlcorrespondingauthor{Matthieu Blanke}{matthieu.blanke@inria.fr}

    \icmlkeywords{Machine Learning, ICML}

    \vskip 0.3in
]



\printAffiliationsAndNotice{}  

\begin{abstract}
   Model-based reinforcement learning is a powerful tool, but collecting data to fit an accurate model of the system can be costly. Exploring an unknown environment in a sample-efficient manner is hence of great importance. However, the complexity of dynamics and the computational limitations of real systems make this task challenging. In this work, we introduce~FLEX, an exploration algorithm for nonlinear dynamics based on optimal experimental design. Our policy maximizes the information of the next step and results in an adaptive exploration algorithm, compatible with generic parametric learning models and requiring minimal resources. We test our method on a number of nonlinear environments covering different settings, including time-varying dynamics. Keeping in mind that exploration is intended to serve an exploitation objective, we also test our algorithm on downstream model-based classical control tasks and compare it to other state-of-the-art model-based and model-free approaches. The performance achieved by~FLEX is competitive and its computational cost is low.
\end{abstract}
\section{Introduction}
\label{section:introduction}

Control theory and model-based reinforcement learning have had a range of achievements in various fields including aeronautics, robotics and energy systems~\cite{Kirk1970,Sutton2018}. For the agent to find an effective control policy, the mathematical model of the environment must faithfully capture the dynamics of the system and thus must be fit with data. However, collecting observations can be expensive: consider for example an aircraft system, for which running experiments costs a lot of energy and time~\cite{gupta1976application}. In this regard, active exploration~(or system identification) aims at exciting the system in order to collect informative data and learn the system globally in a sample-efficient manner~\cite{yang2021exploration}, independent of any control task. Once this task agnostic exploration phase is completed, the learned model can be exploited to solve multiple downstream tasks.

Dynamics may be complex and generally take the form of a nonlinear function of the state.
An example would be air friction, which is essential to consider for an accurate control law, and yet difficult to model from physical principles~\cite{8118153, de2015influence}.
While exploration in linear systems is well understood~\cite{goodwin1977dynamic}, efficiently learning nonlinear dynamics is far more challenging.
For realistic applications, this is compounded by the hard limitations of memory and computational resources of embedded systems~\cite{tassa2012synthesis}. Furthermore, the dynamics or the agent's model of it may vary over time, and the exploration policy must adapt as the data stream is collected. Therefore, it is critical that the algorithm works adaptively, with limited memory storage and that it runs fast enough to be implementable in a real system, while being flexible enough to learn complex dynamics with potentially sophisticated models.


Different approaches have been proposed recently for exploring nonlinear environments, and have proceeded by maximizing an information gain (or uncertainty) on the parameters for specific classes of learning models. An exact computation can be derived for models with linear parametrizations~\cite{pmlr-v100-schultheis20a}, which are however of limited expressivity. Uncertainty can also be computed with Gaussian processes~\cite{buisson2020actively} or approximated using ensembles of neural networks~\cite{pmlr-v97-shyam19a,sekar2020planning} but these approaches suffer from a quadratic memory complexity and an important computational cost respectively, making them unlikely to be implementable in real systems.
In all the above approaches, the inputs are planned episodically by solving a non-convex optimization problem where the nonlinear dynamics are simulated over a potentially large time horizon.
Planning then relies on nonlinear solvers, which may be too slow to run in real time~\cite{kleff2021high}.
Furthermore, planning over large time horizons renders the algorithm unable to adapt to new observations as they are collected. As a result, the agent may spend a long time trusting a wrong model and exploring uninformative states.
Recent works have focused on deriving a fast and adaptive exploration policy for linear dynamics~\cite{greedy}. However, maintaining such guarantees while exploring substantially more complex systems remains a significant challenge and an open area of research.


\paragraph{Contributions}
The present study examines the problem of active exploration of nonlinear environments, with great importance attached to the constraints imposed by real systems.
Based on information theory and optimal experimental design, we define an exploration objective that is valid for generic parametric learning models, encompassing linear models and neural networks.
We derive an online approximation of this objective and introduce FLEX, a fast and adaptive exploration algorithm.
 The sample-efficiency and the adaptivity of our method are demonstrated with experiments on various nonlinear systems including a time-varying environment and the performance of FLEX is compared to several baselines.
We further evaluate our exploration method on downstream exploitation tasks and compare FLEX to model-based and model-free exploration approaches.

\paragraph{Organization} We first introduce the mathematical formalism of our problem in~Section~\ref{section:setting}. In Section~\ref{section:information}, we focus on models with linear parametrizations, for which an information-theoretic objective can be derived. In~Section~\ref{section:online_D-optimal}, we introduce FLEX~(Algorithm~\ref{algorithm:flex}), our exploration algorithm that adaptively maximizes this objective online. In~Section~\ref{section:nonlinear}, we extend our approach to generic, nonlinear models. We test our method experimentally in~Section~\ref{section:experiments}.

\begin{figure*}
    \label{fig:pendulum_exploration}
    \centering
    {\footnotesize $t=1$} \quad
    \raisebox{-.5\height}{\includegraphics[width=.75\linewidth,
            trim={0 1.2cm 0 0cm}
            ,clip]{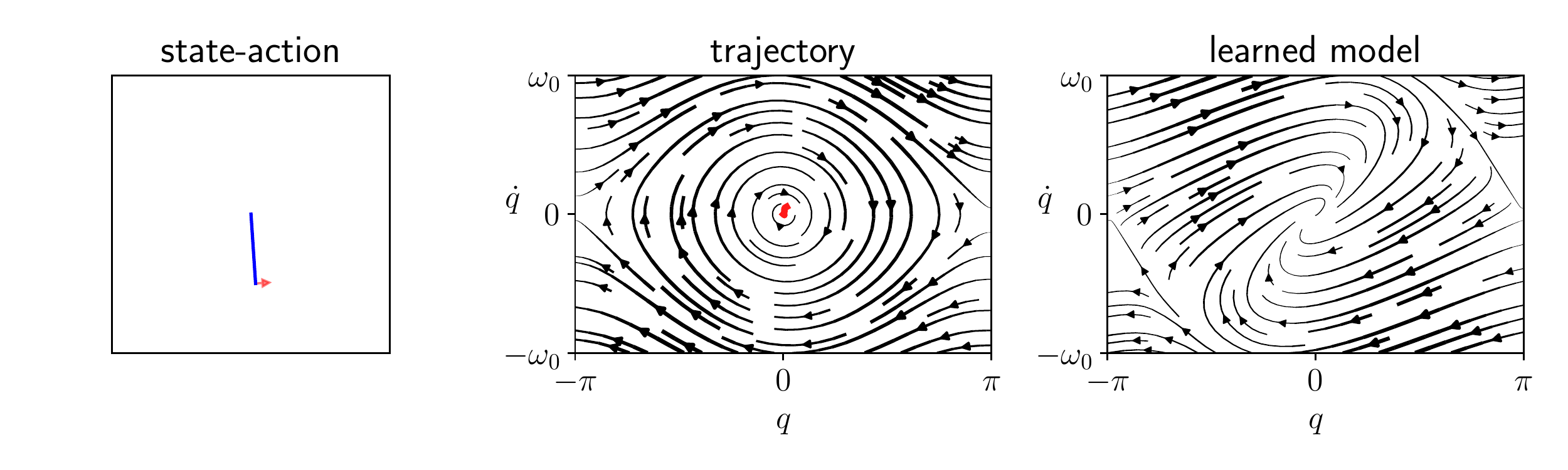}}
    \par
    {\footnotesize $t=100$} \,
    \raisebox{-.5\height}{\includegraphics[width=.75\linewidth,
            trim={0 .5cm 0 1.2cm}
            ,clip]{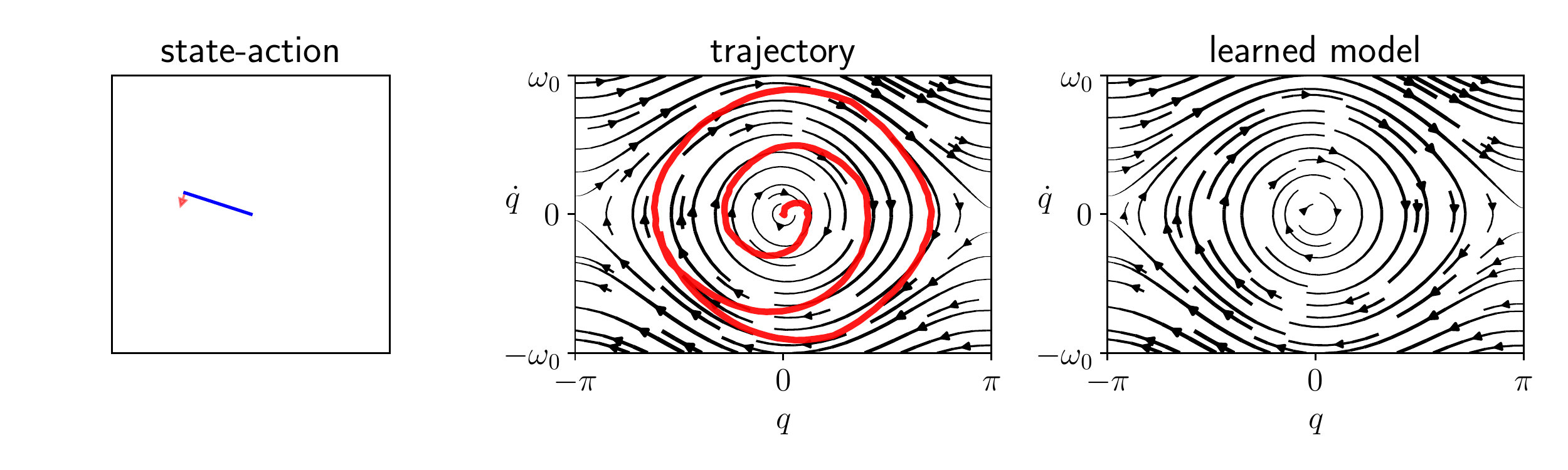}}
    \vskip -0.05in
    \caption{Illustration of active exploration for the dynamics of the pendulum. At each time step, an action is chosen \textbf{(left)}. The decision is taken by using the past trajectory \textbf{(middle)} to learn a model~${f}$ of the dynamics \textbf{(right)} and hence a prediction of the next states.
    }
    \vskip -0.1in
\end{figure*}
\section{Exploring a nonlinear environment}
\label{section:setting}

In nonlinear dynamical systems, the state $x \in \R^d$ and the input $u \in \R^m$ are governed by an equation of the form
\begin{equation}
    \begin{aligned}
        \frac{\ud x}{\ud t}  = f_\star(x, u),
    \end{aligned}
    \label{eq:controlled_dynamics}
\end{equation}
where $f_\star$ is a nonlinear function modeling the dynamics. This function is unknown or partially unknown, and our objective is to learn it from data, with as few samples as possible.
What is observed in practice is a finite number of discrete, noisy observations of the dynamics~\eqref{eq:controlled_dynamics}:
\begin{equation}
    \label{eq:dynamics}
    x_{t+1} = x_t + \ud t f_\star(x_t, u_t) + w_t, \quad 0 \leq t \leq T-1,
\end{equation}
where  $\ud t$ is a known time step, $T$ is the number of observations, $x_t \in \R^{d}$ is the state vector,~${w_t \sim \mathcal{N}(0, \sigma^2 I_d)}$ is a normally distributed isotropic noise with known variance~$\sigma^2$, and the control variables~$u_t \in \R^m$ are chosen by the agent with the constraint~${\Vert u_t \Vert_2 \leq \gamma}$. We assume that~$f$ is a differentiable function.
We write indifferently $f_\star(x,u)$ or~$f_\star(z)$ where~
${z =
            (x \; u)
            \in \R^{d+m}}$ is the state-action pair.
Note that our problem could be formulated in the framework of continuous Markov decision processes~\cite{Sutton2018}, but we find~\eqref{eq:dynamics} more suitable for our approach.

\paragraph{Sequential learning}
The dynamics function~$f_\star$ is learned from past observations with a parametric function~${f}$, whose parameters are gathered in a vector~$\theta \in \R^n$. We denote a generic parametric model by a map~$
    {f}(z, \theta)
$.
At each time~$t$, the observed trajectory yields an estimate~${\theta_{t}= \hat{\theta}(x_{0:t+1}, u_{0:t})}$ following a learning rule~$\hat{\theta}$, such as maximum likelihood. At the end of the exploration, the agent returns a final value~$\theta_T$. Although this learning problem is rich and of an independent interest, we will adopt simple, bounded-memory, online learning rules and focus in this work on the following decision making process.
\paragraph{Sequential decision making}
The agent's decision takes the form of a policy~${\pi : (x_{0:t},u_{0:t-1}|\theta_t)\mapsto u_t}$, mapping the past trajectory to the future input, knowing the current parameter~$\theta_t$.
We denote by~$\Pi_\gamma$ the set of policies satisfying the constraint of amplitude~$\gamma$.
The sequential decision making process is summarized in~Algorithm~\ref{algorithm:exploration} and illustrated in~Figure~\ref{fig:pendulum_exploration}.
The goal of active exploration is to choose inputs that make the trajectory as informative as possible for the estimation of~$f_\star$ with~${f}$, as stated below.
\begin{algorithm}[tb]
    \caption{active exploration}
    \label{algorithm:exploration}
    \begin{algorithmic}
        \STATE \textbf{input} learning model~${f}$, time horizon~$T$, time step $\ud t$, policy~$\pi \in \Pi_\gamma$, learning rule~$\hat{\theta}$
        \STATE \textbf{output} parameter estimate ${\theta_{T}}$
        \FOR{$0 \leq t \leq T-1$}
        \STATE  choose $u_t = \pi(x_{0:t}, u_{0:t-1} | \theta_t)$
        \STATE  observe $x_{t+1} = x_t + \ud t f_\star(x_t, u_t) + w_t$
        \STATE  update $\theta_{t+1}= \hat{\theta}(x_{0:t+1}, u_{0:t+1})$
        \ENDFOR
    \end{algorithmic}
\end{algorithm}
\vskip -3in

\paragraph{The problem}  For an arbitrary learning model of the dynamics~${f}$ provided with a learning rule~$\hat{\theta}$ and for a fixed number of observations~$T$, the goal is to find an exploration policy~$\pi$ for which the learned model is as close to~$f_\star$ as possible at the end of exploration. Formally, we define the estimation error of parameter~$\theta$ as
\begin{equation}
    \label{eq:performance_evaluation}
    \varepsilon(\theta) = \Vert {f}(., \theta) - f_\star \Vert_{L^2},
\end{equation}
\vskip -0.1in
and look for a policy yielding the best estimate of~$f_\star$ at time~$T$:
\begin{equation}
    \label{eq:exploration_objective}
    \underset{\pi \in \Pi_\gamma}{\min} \quad \mathbb{E} [\varepsilon(\theta_T) | \pi],
\end{equation}
\vskip -0.15in
where the expectation is taken over the stochastic dynamics~\eqref{eq:controlled_dynamics} and possibly the randomness induced by the policy.
\paragraph{Practical considerations}
In addition to the sample efficiency objective~\eqref{eq:exploration_objective}, we attach great importance to the three following practical points that emerge from the observations of Section~\ref{section:introduction}. \textit{Adaptivity}: as opposed to episodic planning for which~$\pi(.|\theta)$ remains constant with respect to~$\theta$ throughout long time intervals, we want our policy to accommodate to new observations at each time step. \textit{Computational efficiency}: evaluating the policy~$\pi$ should require limited computational resources. \textit{Flexibility}: our policy should be valid for a broad class of models~${f}$, as we will see with the following examples.
\begin{example}[Damped pendulum]
    \label{example:pendulum}
    The angle $q$ of a pendulum driven by a torque $u$ satisfies the following nonlinear differential equation:
    \begin{equation}
        \label{eq:pendulum-dynamics}
        \ddot{q} + \alpha \dot{q} + \omega_0^2 \sin q = b u.
    \end{equation}
    This second-order system can be described by the bidimensional state variable $x = (q, \dot{q})\transp{} \in \R^2$ and the nonlinear map $f_\star(q, \dot{q}, u) = (\dot{q}, -\alpha \dot{q} - \omega_0^2 \sin q + bu)\transp{}$.
\end{example}
\begin{example}[Nonlinear friction]
    \label{example:friction}
    Using the same notations, the dynamics of a mass subject to a control force and friction is given (in the case of one-dimensional system for simplicity) by~
    ${f_\star(q, \dot{q}, u) = (
                \dot{q}, \mathrm{friction}(q, \dot{q}) + bu)\transp{}}$.
    The friction force is notoriously difficult to model. One possibility would be the nonlinear function~${\mathrm{friction}(q, \dot{q}) = -\alpha |\dot{q}| q}$~\cite{zhang2014quadrotor}.
\end{example}
The choice of  model is crucial and depends on prior knowledge about the system. In~Example~\ref{example:pendulum}, if the system is known down to its scalar parameters, learning the dynamics reduces to linear regression. In the opposite case where little or nothing is known about the structure, as in~Example~\ref{example:friction}, it would be desirable to fit more sophisticated parametric models. For instance, the recent successes of neural networks and their ability to express complex functions and to be trained online make them a promising option for our exploration task~\cite{goodfellow2016deep}.

Importantly, the exploration policy should depend on the learned model and drive the system to regions with high uncertainty.
In contrast, random exploration fails to explore nonlinear systems globally: typically, friction forces pull the system toward a fixed point, while learning the environment requires large amplitude trajectories and hence temporal coherence in the excitation.

\section{Information-theoretic view of exploration}
\label{section:information}

Given a model of the dynamics, how do we choose inputs that efficiently navigate in phase space towards informative states~? This choice should be guided by some measure of the information that the trajectory provides about our learning model. In this section, we turn to information theory and study the simplest form of models: linear models provide us not only with a natural learning rule but also with an information-theoretic measure of exploration.
We then leverage this criterion to define an optimization objective.

\subsection{Linear models}
\label{section:linear_model}

An important class of models is the class of functions with an affine dependence on  the parameters. In the remainder of this work, we use the word ``linear" for an affine dependence. Note that the term ``linear" refers to the parameter dependence of a model, but the dependence in the state-action pair~$z$ is still assumed to be nonlinear in general.
\begin{definition}[Linear model]
    \label{definition:linear-model}
    The most general form for a linear dependence of~${f}$ in its parameters  is
    \begin{equation}
        \label{eq:linear-model}
        {f}(z, \theta) = V(z) \times\theta + c(z),
    \end{equation}
    where the features~$V(z) \in \R^{d\times n}$
    and~$c(z) \in \R^d$
    are independent of~$\theta$. We denote by~${v^{(j)} \in \R^n, 1 \leq j \leq d}$ the rows of~$V$, and we define~${V_t := V(z_t)}$.
\end{definition}
The class of linear models is critically important for several reasons.
First, the dynamics can very often be formulated as an affine function of some well-defined scalar parameters, hence allowing for efficient estimation by ordinary least squares~\cite{underactuated}. Besides, a natural mathematical exploration objective from optimal experimental design theory can be derived for linear models (see~Section~\ref{section:optimal-design}). Finally, the computations allowed for linear models can be generalized to nonlinear models, as we will see in~Section~\ref{section:nonlinear}.

\begin{sloppypar}
    \begin{example}[Learning the pendulum]
        \label{example:learning-pendulum}
        When the pendulum of~Example~\ref{example:pendulum} is known up to the parameters~${\theta =
                    (\omega_0^2, \alpha, b) \transp{}}$
        then the dynamics can be learned with a linear model as defined in~\eqref{eq:linear-model} with~$n=3$ by defining
        \begin{equation}
            V(z) = \begin{pmatrix}
                0          & 0           & 0 \\
                -\sin \phi & -\dot{\phi} & u
            \end{pmatrix}, \quad
            c(z) = (\dot{\phi}, 0) \transp{}.
        \end{equation}
        The real dynamics function is~${f_\star(z) = V(z)\times \theta_\star} + c(z)$ with~$\theta_\star$ the true parameters.
    \end{example}
\end{sloppypar}
An important case of linear models is when the parameters are separated in a matrix form according to row-wise dependence, as follows.
\begin{sloppypar}
    \begin{definition}[Matrix model]
        \label{definition:matrix-model}
        We call a matrix model a parametrization of the form
        \begin{equation}
            \label{eq:matrix-model}
            {f}(z, \Theta) = \Theta \times \phi(z),
        \end{equation}
        where~${\phi : \R^{d+m} \rightarrow\R^{n'}}$ is a feature map and~${\Theta \in \R^{d\times r}}$ is a parameter matrix.
        The structure~\eqref{eq:linear-model} is recovered by defining~$\theta$ as the vectorization of~$\Theta$ of size~$n= d\times n'$, and the features as the block matrix~${V = \mathrm{diag}(\transp{\phi}, \dots, \transp{\phi}) \in \R^{d\times n}}$.
    \end{definition}
\end{sloppypar}
\begin{sloppypar}
    \begin{example}[Linear dynamics]
        \label{example:LTI-dynamics}
        Consider the case of a linear time-invariant system:~${f_\star(x, u) = A_\star x + B_\star u}$. A natural parametrization is~${{f}(z, \Theta) = \Theta \times z}$,
        with the parameters~${\Theta = (A \, B)} \in \R^{d\times (d+m)}$.
        This is a matrix model~\eqref{eq:matrix-model}
        where~$\phi(z) = z$ and $n'=d+m$.
    \end{example}
\end{sloppypar}
\begin{example}[Random Fourier Features]
    In~\cite{pmlr-v100-schultheis20a}, the nonlinear dynamics are modeled with a Random Fourier Features model, which takes the form ~\eqref{eq:matrix-model}, with~$\phi(z)$ a random feature.
\end{example}



\subsection{Optimal experimental design}
\label{section:optimal-design}

Let~$y$ be an observation whose distribution~$p(y|\pi, \theta)$ depends on a parameter~$\theta \in \R^n$ and a decision variable (or design) denoted~$\pi$. Experimental design theory provides a quantitative answer to the question: how informative are the observations~$y$ for estimating~$\theta$ when the decision~$\pi$ is taken?
The information is defined as a scalar functions of the Fisher information matrix~\cite{fedorov2010optimal} as follows.
\begin{definition}[Fisher information and information gain]
    We denote~$\ell(y, \theta) = \log p(y|\theta)$ the log-likelihood of the distribution.
    The observed Fisher information matrix~\cite{gelmanbda04} of observation~$y$ at some parameter value~$\theta$ is
    \begin{equation}
        I(y, \theta) = - \frac{\partial^2 \ell}{\partial \theta^2}(y, \theta)  \quad \in \R^{n \times n}.
    \end{equation}
    The~D-optimal information gain is defined as
    \begin{equation}
        g(\pi | \theta) = \log \det \left(\mathbb{E}\left[I(y, \theta) \big\vert \pi, \theta \right]\right).
    \end{equation}
\end{definition}
\begin{example}[Linear regression]
    In a linear regression,~${\pi \in \R^d}$ is the regressor and $y=\transp{\pi}\theta + w \in \R^d$ with the noise~${w\sim \mathcal{N}(0, I_d)}$, yielding~$I(y, \theta) = \pi \transp{\pi} \in \R^{d\times d}$ and $g(\pi) = \log \det \pi \transp{\pi}$.
\end{example}
This information gain quantifies the information about~$\theta$ provided by the observations with design~$\pi$, and it may be interpreted as the volume of the confidence ellipsoid for the parameter vector~$\theta$.
Several other functionals can be used instead of~$\log \det$, leading to other optimality criteria. The D-optimality criterion
benefits from a property of scale invariance~\cite{pukelsheim2006optimal} and
is suitable for our online setting because it provides a simple rank-one update formula, as we will see in~Section~\ref{section:online_D-optimal}.
A D-optimal design maximizes the information gain:
\begin{equation}
    \label{problem:optimal-design}
    \underset{\pi}{{\max}} \quad g(\pi | \theta).
    \vspace{-0.05in}
\end{equation}
From a Bayesian perspective, D-optimality minimizes
the entropy of the expected posterior on~$\theta$, or equivalently the mutual information between the current prior and the expected posterior~\cite{Chaloner1995}.

\subsection{Optimally informative inputs for exploration}
In this section, we apply the optimal experimental design framework to our dynamical setting, where the Gaussian noise assumption and the linear structure of the model allow for exact computations. The observations are the trajectory~${y=(x_{0:t}, u_{0:t-1})}$ and the design is the policy~$\pi$ generating the inputs~$u_{0:t-1}$.
\begin{definition}[Gram matrix]
    \label{definition:gram}
    An important quantity is the Gram matrix of the features, defined as
    \begin{equation}
        \label{eq:gram}
        \begin{aligned}
            M_t
             & =  \sum\limits_{s=0}^{t-1}\transp{V_s}V_s \quad \in \R^{n\times n}.
        \end{aligned}
    \end{equation}
\end{definition}
\begin{proposition}
    \label{proposition:fisher_dynamics}
    Assume a linear model~\eqref{eq:linear-model} for the data-generating distribution of the trajectory, with~$c=0$ for simplicity. Then, the maximum likelihood estimate of~$\theta$ is
    \begin{equation}
        \label{eq:ols}
        \hat{\theta}(y) = M_t^{-1} \sum\limits_{s=0}^{t-1} \transp{V_s}x_{s+1}
    \end{equation}
    and the Fisher information matrix is
    \begin{equation}
        \label{eq:fisher_exploration}
        I(y, \theta) = \frac{1}{2 \sigma^2} M_t.
    \end{equation}
    \vskip -1in
\end{proposition}
Note that $I$ does not depend explicitly on~$\theta$ but only on the observations, because the model is linear in the parameters.

It follows from~\eqref{problem:optimal-design} and~Proposition~\ref{proposition:fisher_dynamics} that, in our setting, D-optimal inputs solve the following optimal control problem:
\begin{equation}
    \label{problem:D_optimal_optimal_control}
    \begin{aligned}
        \underset{(z_t)}{\text{maximize}} \quad & \log \det \left(
        \sum\limits_{t=0}^{T-1}  \transp{V_t}V_t \right)
        \\
        \text{subject to} \quad                 & x_{t+1} = x_t + \ud t  f(x_t, u_t),
        \\
                                                & \Vert u_t \Vert_2 \leq \gamma^2, \quad 0 \leq t \leq T-1,
    \end{aligned}
\end{equation}
where we neglect the noise in the dynamics for simplicity, and recall that~$V_t = V(z_t)$.
\begin{remark}[Optimal design for matrix models]
    \label{remark:optimal-design_feature-models}
    For models of the form~\eqref{eq:matrix-model}, one can readily show that an equivalent objective is obtained by defining~$V_t := \transp{\phi(z_t)}$ instead of~$V_t=V(z_t)$.
\end{remark}
\subsection{Sequential learning}
Linear models can be learned online with the recursive least squares formula for the estimator~\eqref{eq:ols}. Assuming for simplicity~$d=1$,~$c=0$ and denoting~${v_t=\transp{V(z_t)} \in \R^n}$, online learning takes the form
\begin{subequations}
    \begin{align}
        \theta_{t+1} & = \theta_t - M_t^{-1}v_t(\transp{v_t}\theta_t - x_{t+1})
        \\
        \label{eq:ols_gradient}
                     & = \theta_t - H_t \nabla \ell_t(\theta_t).
    \end{align}
\end{subequations}
\todo{\color{blue} remove the gradient}
with the squared error loss~
$\ell_t(\theta) = \frac{1}{2} \times \Vert  \transp{v_t} \theta - x_{t+1} \Vert^2_2$ and the matrix step size~$H_t:= M_t^{-1}$.
Equation~\eqref{eq:ols_gradient} makes it transparent that recursive least squares is an online gradient descent step.
The memory cost of learning is~$\mathcal{O}(n^2)$ for the storage of~$M_t$ and~$\theta_t$. For $d > 1$, there is one update for each row, hence~$d$ updates per time step $t$.

\section{Adaptive D-optimal exploration with FLEX}
\label{section:online_D-optimal}
We adopt D-optimality~\eqref{problem:D_optimal_optimal_control} as an objective for our exploration policy.
However, this problem is non-convex and providing a numerical solution is computationally challenging. Furthermore, the dynamics constraint is unknown and can only be approximated with the current knowledge of the dynamics, which is improved at each time step. Although previous approaches have opted for an episodic optimization with large time horizons~\cite{wagenmaker2021taskoptimal,pmlr-v100-schultheis20a}, it is desirable to  update the choice of inputs at the same frequency as they are collected. In this section, we introduce~FLEX~(Algorithm~\ref{algorithm:flex}), an adaptive D-optimal exploration algorithm with low computational complexity.
In this section, we use the notations of linear models as defined in~Section~\ref{section:information}.
We will show how this formalism extends to arbitrary, nonlinear models in~Section~\ref{section:nonlinear}.

\subsection{One-step-ahead information gain}

Since we seek minimal complexity and adaptivity, we choose to devote the computational effort at time~$t$ to the choice of the next input~$u_t$ only.
We want to define an informativeness measure~$F_t$ for input~$u_t$ and solve a sequence of problems of the form
\vskip -0.1in
\begin{equation}
    \label{problem:planning_u}
    \begin{aligned}
        u_t \in \quad & \underset{u \in \R^m}{\text{argmax}} \quad
        F_t(u)
        \\
                      & \text{subject to} \quad                         \Vert u \Vert_2 \leq \gamma^2.
    \end{aligned}
\end{equation}
\vskip -0.1in
As stated before, we attach great importance to the computational time of solving~\eqref{problem:planning_u}.

The function~$F_t$ should quantify the information brought by~$u_t$, for which we derived a mathematical expression in our information-theoretic considerations of~Section~\ref{section:information}.
Therefore, we want the problem sequence~\eqref{problem:planning_u} to be an approximation of the problem~\eqref{problem:D_optimal_optimal_control}.

A greedy approximation can be derived as follows.
At time~$t$, the past trajectory~$z_{0:t}$ is known and the choice of~$u_t$ immediately determines the next state~$x_{t+1}$. We define~$F_t$ as the predicted information gain truncated at~$t+1$.
\begin{definition}[Predicted information gain]
    Letting~${x(u) := x_t + \ud t {f}(x_t, u, \theta_t)}$ be the one-step-ahead state prediction and~${z(u) := (x(u), 0)}$, we define
    \begin{equation}
        \label{eq:one-step_objective}
        F_t(u) = G(z(u))
        \vspace{-.2cm}
    \end{equation}
    with
    \begin{equation}
        \label{eq:one-step_information}
        G(z)=\log \det ( M_t + \transp{V(z)}V(z)).
    \end{equation}
\end{definition}
\vspace{-.1cm}
When~$V$ is of rank one, a simpler formula can be derived.
\begin{lemma}[Determinant Lemma]
    \label{lemma:determinant}
    By choosing a row~${v:=v^{(k)}}$ for~$1 \leq k \leq d$ of the feature matrix~$V$ and approximating~
    ${\transp{V}V \simeq v \transp{v}}$,
    the expression for the information gain can be simplified as follows:
    \begin{equation}
        \label{eq:D-optimal_rank-one}
        G(z) = \log \det M_t +  \transp{v}(z) M_t^{-1} v(z).
    \end{equation}
\end{lemma}
\vskip -0.1in
We adopt the rank-one approximation of~Lemma~\ref{lemma:determinant} in the remainder of this work.
Note that the index~$k$ of row~$v$ can be either drawn randomly or chosen using prior knowledge: the~$k$-th row of~$V$ is informative if the~$k$-th component of the model is sensitive with respect to the parameters. One could also design a numerical criterion for this choice.

\subsection{Computing D-optimal inputs}

For linear systems, recent approaches have shown that a greedy D-optimal policy yields good exploration performance~\cite{greedy}.
However for nonlinear systems, the nonlinearity of~$v(z)$ makes the maximization problem~\eqref{problem:planning_u} challenging.
To obtain a simpler optimization problem, we linearize the model with respect to the state. Intuitively, since we are planning between~$t$ and~$t + \ud t$, the corresponding change in the state is small so it is reasonable to use a linear approximation to the mapping~$z \mapsto v(z)$.
\begin{proposition}
    \label{proposition:quadratic}
    Linearizing our objective~\eqref{eq:one-step_objective} to first order in~$ \ud t$ yields the following approximation to the optimization problem~\eqref{problem:planning_u}:
    \begin{equation}
        \begin{aligned}
            \label{eq:quadratic-problem}
            \underset{u \in \mathbb{R}^m}{\text{maximize}} \quad & \transp{u}Qu - 2 \transp{b}u
            \\
            \text{subject to} \quad                              & \Vert u \Vert_2^2 \leq \gamma^2,
        \end{aligned}
    \end{equation}
    where $Q$ and $b$ are computed in terms of the Gram matrix~${M:=M_t \in \R^{n \times n}}$ and the vector~$v$ and the derivatives~${D := \partial v / \partial x \in \R^{n \times d}}$, and ${B := \ud t \, \partial {f} / \partial u \in \R^{d \times m}}$ evaluated at $\bar{z} := z(u=0)$, as follows
    \begin{equation}
        \label{eq:matrices}
        \begin{aligned}
            Q = \transp{B} \transp{D} {M}^{-1} {D} B  \quad \in \mathbb{R}^{m \times m},
            \\
            b = -\transp{B} \transp{D}{M}^{-1}v \quad \in \mathbb{R}^{m}.
        \end{aligned}
    \end{equation}
\end{proposition}
\begin{remark}[Linear dynamics]
    \begin{sloppypar}
        When the dynamics are linear as in~Example~\ref{example:LTI-dynamics}, it follows from Example~\ref{example:LTI-dynamics} and~Remark~\ref{remark:optimal-design_feature-models} that~${v(z) = z}$, hence~${\partial v / \partial x = I_{n,d}}$ and~${M_t = \sum_{s=0}^{t-1} z_s \transp{z_s} \in \R^{d \times d}}$. Then, applying Proposition~\ref{proposition:quadratic} yields
        ~${Q = \transp{B} M_t B}$ and~${b = \transp{B}M_t(I_d + \ud t A) x_t}$ and we find exactly the greedy optimal design algorithm of~\cite{greedy}.
    \end{sloppypar}
\end{remark}
Our exploration policy is summarized in~Algorithm~\ref{algorithm:flex}. Note that since it optimizes a greedy objective at each time step, it is adaptive by nature and it does not require the knowledge of the time horizon~$T$.
The following result shows that solving~\eqref{eq:quadratic-problem} can be achieved at low cost, ensuring that our policy is computationally efficient.
\begin{proposition}
    \label{proposition:numerical_solution}
    Problem~\eqref{eq:quadratic-problem} can be solved numerically at the cost of a scalar root-finding and a~$m \times m$ matrix eigenvalue decomposition.
\end{proposition}
\begin{algorithm}[tb]
    \caption{{F}ast {L}inearized {EX}ploration (FLEX)}
    \label{algorithm:flex}
    \begin{algorithmic}
        \STATE \textbf{input} model ${f}$, horizon $T$, time step $\ud t$, first estimate~$\theta_0$
        \STATE \textbf{output} parameter estimate $\theta_T$
        \FOR{$0 \leq t \leq T-1$}
        \STATE  compute $Q_t$, $b_t$ \quad from \eqref{eq:matrices}
        \STATE  choose  $u_t \in \underset{
                \transp{u}u \leq \gamma^2
            }{\mathrm{argmax}} \; \transp{u}Q_t u - 2 \transp{b}_t u$ (Proposition~\ref{proposition:quadratic})
        \STATE  observe $x_{t+1} = x_t + \ud t \, f_\star(x_t, u_t) + w_t$
        \STATE  compute $\ell_t(\theta) = \frac{1}{2}\Vert {f}(x_t, u_t, \theta) - (x_{t+1} - x_t) / \ud t \Vert^2_2$
        \STATE  update $\theta_{t+1} = \theta_t - H_t \nabla \ell_t(\theta_t)$ \quad as in \eqref{eq:online-gd}
        \ENDFOR
    \end{algorithmic}
\end{algorithm}

\section{From linear models to nonlinear models}
\label{section:nonlinear}

For the cases when no prior information is available about the structure of the dynamics (see~Example~\ref{example:friction}), it is desirable to generalize the policy derived in~Section~\ref{section:online_D-optimal} to more complex models that are not linear in the parameters.
In this section, we extend ~Algorithm~\ref{algorithm:flex} to generic, nonlinear parametric models. We assume that~${f}$ is doubly differentiable with respect to~$z$ and~$\theta$. 

\subsection{Linearized model}
\label{section:linearized-OD}

The developments of~Section~\ref{section:online_D-optimal} are based on the linear dependence of the model on the parameters. For nonlinear models, a natural idea is to make a linear expansion of the model: assuming that the parameter vector is close to a convergence value~$\theta_\star$, we can linearize ${f}$ to first order in~${\theta-\theta_\star}$:
\vskip -.25in
\begin{equation}
    \label{eq:parameter_linearization}
    {{f}(z, \theta) \simeq {f}(z, \theta_\star) +\frac{\partial {f}}{\partial \theta}(z, \theta_\star) \times (\theta - \theta_\star)}.
    \vspace{-.15cm}
\end{equation}
In the limiting regime where this linear approximation would hold,~${f}$ would be a linear model with features
    \begin{equation}
        \label{eq:jacobian_feature}
        V(z) = \frac{\partial {f}}{\partial \theta}(z,\theta_\star) \quad \in \R^{d\times n}.
    \vspace{-.4cm}
    \end{equation}

Considering this analogy, we can 
generalize D-optimal experimental design~\cite{mackay1992information} and we want to extend the results of~Section~\ref{section:online_D-optimal} to nonlinear models. 

\subsection{Online exploration with nonlinear models}
\label{section:online-exploration_nonlinear}

In our dynamical framework, we expect the approximation~\eqref{eq:parameter_linearization} to be increasingly accurate as more observations are collected, hence motivating the generalization of the exploration strategy developed in~Section~\ref{section:online_D-optimal} to nonlinear models. The features of the linearized model~\eqref{eq:parameter_linearization} are unknown because the Jacobian~\eqref{eq:jacobian_feature} depends on the unknown parameter~$\theta_\star$ in general. However, we can approximate~$\theta_\star$ by the current estimate~$\theta_t$ at each time step along the trajectory. By defining
\vskip -0.3cm
\begin{equation}
    \label{eq:regressor_nonlinear}
    V_t := \frac{\partial {f}}{\partial \theta}(z_t, \theta_t),
\end{equation}
we extend the notion of the Gram matrix in~Definition~\ref{definition:gram} as well as the optimal control problem~\eqref{problem:D_optimal_optimal_control} to nonlinear models.
Note that since~$V_t$ depends only on~$(z_t, \theta_t)$, the Gram matrix can still be computed online along the trajectory.
Similarly, we extend the approach developed in~Section~\ref{section:online_D-optimal}
by defining
\begin{equation}
    \label{eq:predicted-feature_nonlinear}
    v(z) := \nabla_\theta {f}^{(k)}(z, \theta_t)
\end{equation}
with~$1\leq k \leq d$ chosen as in~Lemma~\ref{lemma:determinant}.
With these quantities defined,~Algorithm~\ref{algorithm:flex} is extended to arbitrary  models.

\begin{remark}[Consistency with linear models]
    When the model is exactly linear in the parameter as in~\eqref{eq:linear-model}, the Jacobian is ~$\partial {f} / \partial \theta = V(z)$ so~\eqref{eq:regressor_nonlinear} is consistent with the definition of~$V_t$ in~Definition~\ref{definition:linear-model}.
        Therefore,~\eqref{eq:regressor_nonlinear} can be viewed as a generalization of the optimal experimental design exploration of~Section~\ref{section:online_D-optimal} to nonlinear models.
    \end{remark}


\begin{figure*}[bt]
    \centering
    \includegraphics[width=.22\linewidth,
        trim={0 .5cm 0 0},clip
    ]{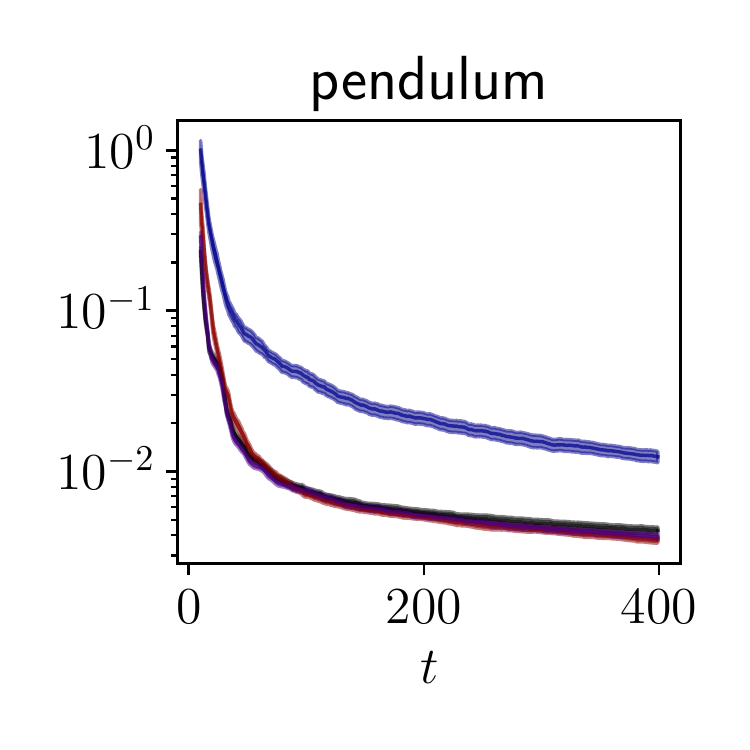}
    \includegraphics[width=.22\linewidth,
        trim={0 .5cm 0 0},clip
    ]{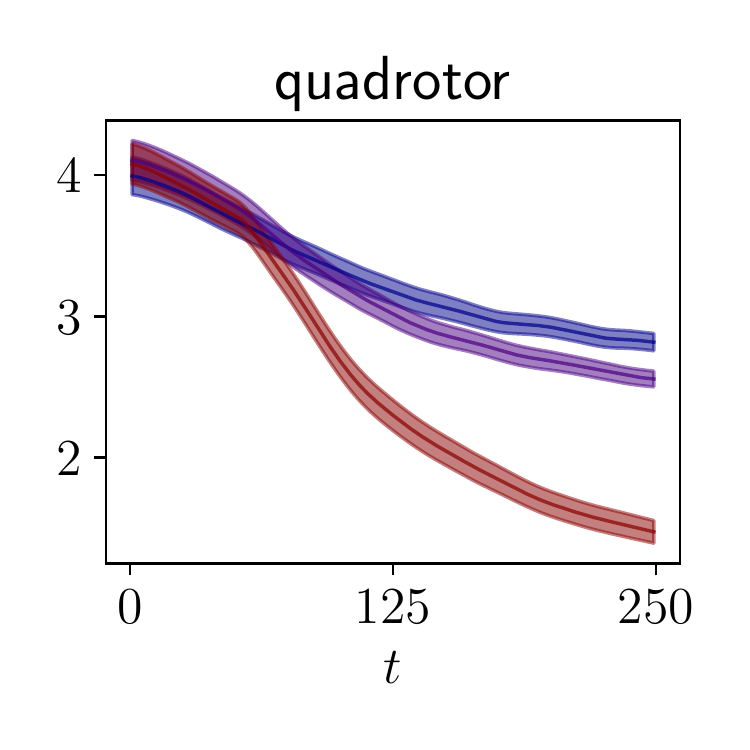}
    \includegraphics[width=.22\linewidth,
        trim={0 .5cm 0 0},clip
    ]{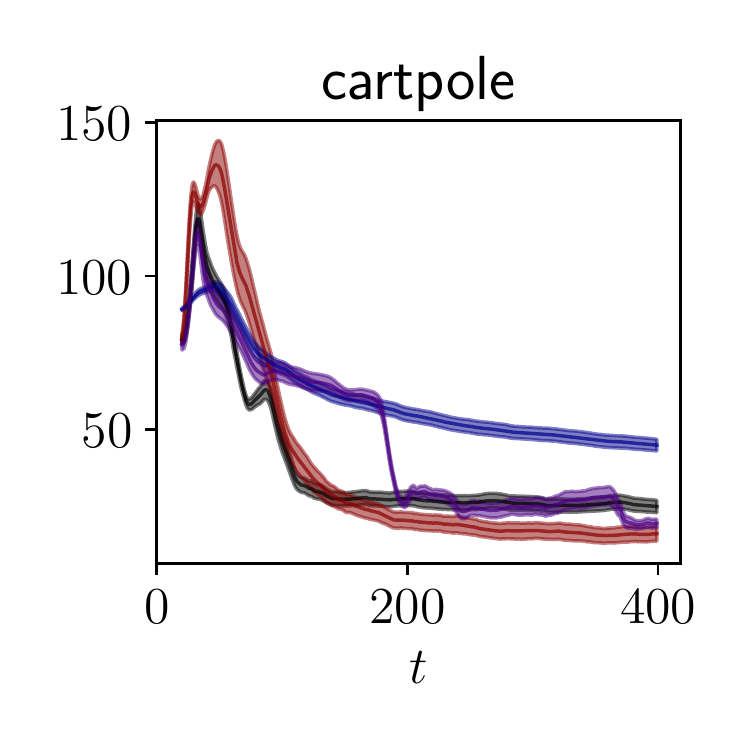}
    \includegraphics[width=.22\linewidth,
        trim={0 .5cm 0 0},clip
    ]{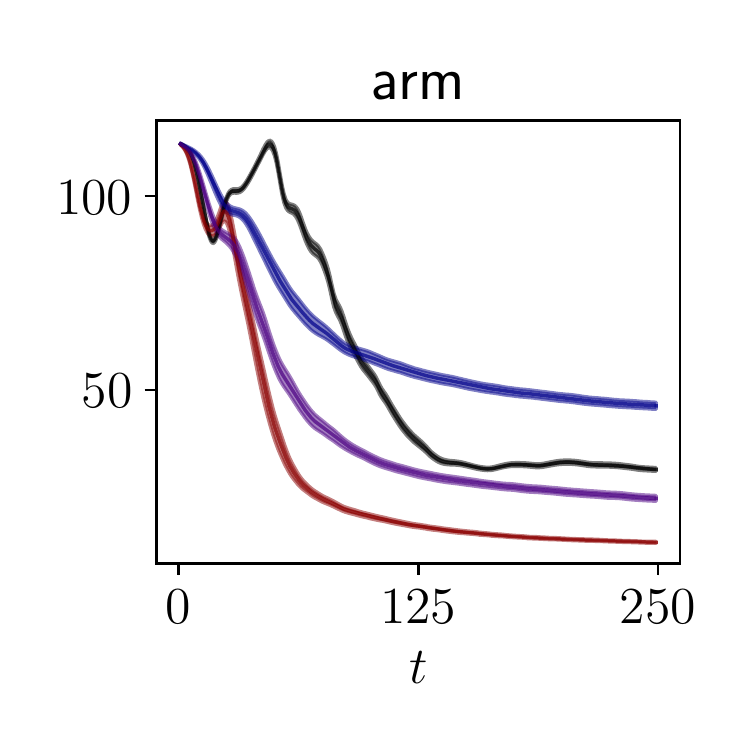}
    \raisebox{.05\height}{
        \includegraphics[height=3.2cm]{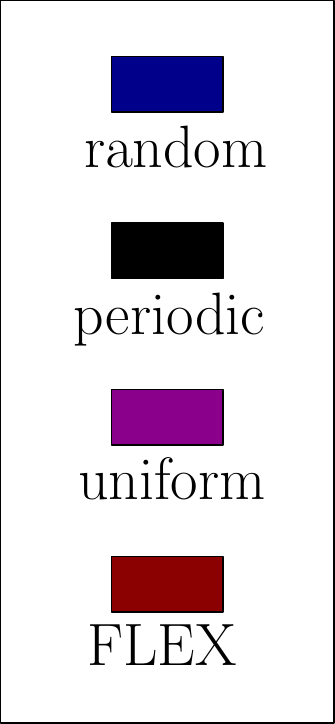}
    }
    \vskip -0.1in
    \caption{Evaluation error over time for different environments as a function of time $t$ averaged over 100 trials.}
    \vskip -0.1in

    \label{figure:benchmark}
\end{figure*}
\subsection{Computational perspective}

In~Algorithm~\ref{algorithm:flex}, we need to compute~$D=\partial v / \partial x$, which amounts to computing the derivatives of~$f$ in both~$\theta$ and~$z$:
\begin{equation}
    \frac{\partial v}{\partial x} = \left( \frac{\partial^2 {f}^{(k)}}{\partial  x_j \partial \theta_i}\right)_{
    \substack{1 \leq i \leq n \\ 1 \leq j \leq d}
    }  \in \R^{n \times d}.
\end{equation}
For neural networks, this matrix can be computed by automatic differentiation. The cost of solving~\eqref{eq:quadratic-problem} does not depend on~$n$ so the computation of~$D$ becomes the computational bottleneck for large models. The complexity of the latter operation is~$\mathcal{O}(nd)$ with automatic differentiation.
\subsection{Sequential learning}
We train nonlinear models using online gradient descent:
\begin{equation}
    \label{eq:online-gd}
    \begin{aligned}
        \theta_{t+1} = \theta_t - H_t \nabla \ell_t(\theta_t),
    \end{aligned}
\end{equation}
which extends~\eqref{eq:ols_gradient}. For neural networks, online learning with an adaptive learning rate (and scalar~$H_t)$ is known to be effective~\cite{bottou2012stochastic, 2015-kingma}. The gradient step can be averaged over a batch for smoother learning, at the cost of storing a small amount of data points.


\section{Experiments}
\label{section:experiments}
We run several experiments to validate our method.
Our code and a demonstration video are available at~{\footnotesize \url{https://github.com/MB-29/exploration}}. More details about the experiments can be found in~Appendix~\ref{appendix:details}.

\subsection{Exploration benchmark}
\label{section:benchmark}

We first test our policy on various nonlinear environments from classical control, covering different values for~$d$ and~$m$. We compare its performance in terms of sample efficiency to that of various baselines. The agents have the same learning model, but different exploration policies. Random exploration draws inputs at random. For pendulum-like environments, a periodic oracle baseline excites the system at an eigenmode, yielding resonant trajectories of large amplitude. A baseline called ``uniform" maximizes the distance of the trajectory points in the state space: it optimizes objective~\eqref{eq:one-step_objective} with~${{G(z) = \frac{1}{2}\sum_{s=0}^{t} \Vert {x}- x_s \Vert^2}}$.

\paragraph{Experimental setup}
 The learning models include various degrees of prior knowledge on the dynamics. A linear model is used for the pendulum, and neural networks are used for the other environments.
 The Jacobians of~Proposition~\ref{proposition:quadratic} are computed using automatic differentiation.
 At each time step, the model is evaluated with~\eqref{eq:performance_evaluation} computed over a fixed grid.

\begin{figure}
    \centering
    \includegraphics[width=.47\linewidth,
    ]{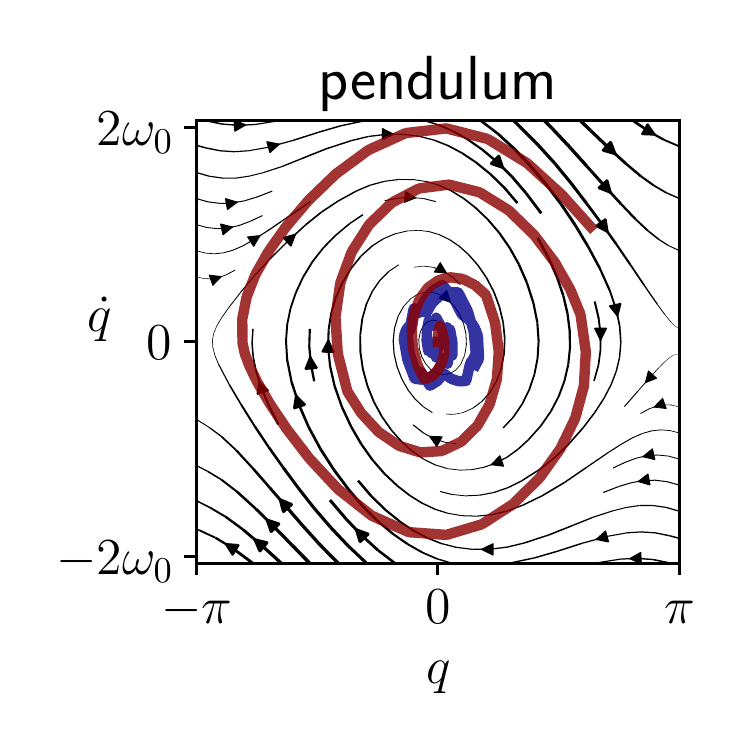}
    \includegraphics[width=.47\linewidth,
    ]{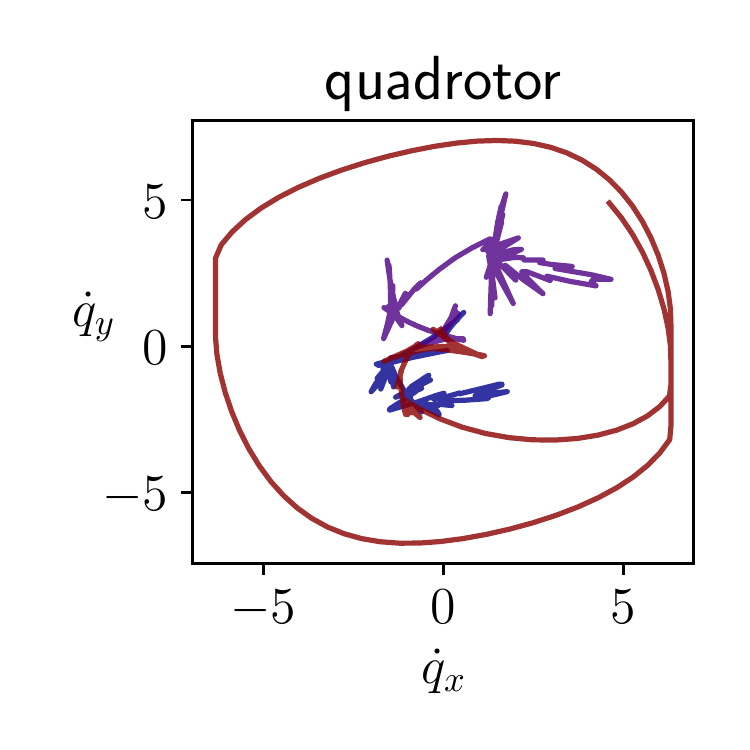}
    \vskip -0.2in
    \caption{
        Trajectories in phase space.  
        }
        \label{figure:trajectories}
        \vskip -0.1in
\end{figure}

\paragraph{Results}
The results are presented in~Figure~\ref{figure:benchmark}. Our algorithm is sample-efficient and it outperforms the baselines in all the environments.
We also display the trajectories obtained with our policy in phase space in~Figure~\ref{figure:trajectories}. 
Not only does FLEX produce informative trajectories of large amplitude, but more specifically it devotes energy so as to explore regions with higher uncertainty, unlike the baselines. Although the policy optimizes the information in a greedy fashion, it interestingly produces inputs with long-term temporal coherence. This is illustrated in our demonstration video. High-dimensional exploration is tackled in~Appendix~\ref{appendix:chain}.

\subsection{Tracking of time-varying dynamics}
\begin{figure*}[bt]
    \includegraphics[width=.925\linewidth]{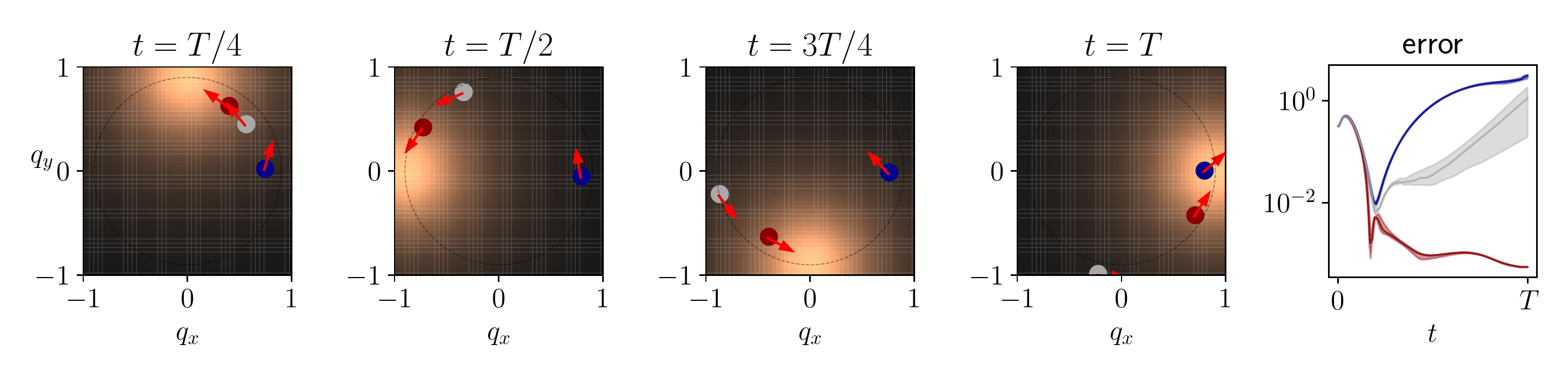}
    \raisebox{.4\height}{
        \includegraphics[width=.056\linewidth]{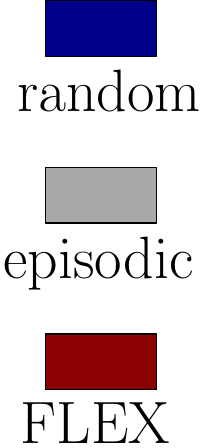}
    }
    \vskip -0.2in
    \caption{Time-varying force field system. \textbf{Left.} Trajectories of the agents in the position space~$(q_x,q_y)$. The color gradient represents the force field and the red arrows represent the actions of the agents. \textbf{Right.} The error curves over time, averaged over 100 trials.}
    \label{figure:adaptive}
\end{figure*}
In real systems, dynamics may vary over time and an adaptive exploration is crucial for accommodating to changes in the environment. We test the adaptivity of our method by exploring a time-varying system, and compare to an episodic agent. 
The system is a central, repulsive force field centered on a star, moving around a circle uniformly at a period~$T$. The agent is a spaceship and the control variable is the acceleration. The force field varies significantly only in the vicinity of the star: the spaceship learns information about the dynamics only when it is close, and hence needs to track the star.

\paragraph{Experimental setup}
The model learns the center and the radius of the force field, yielding a nonlinear parametrization~(see~Appendix~\ref{appendix:time-varying}). With this model, three agents learn the dynamics: a random policy, FLEX, and an episodic agent that plans D-optimal inputs by solving~\eqref{problem:D_optimal_optimal_control} over a time horizon of~$T/20=50$ repeatedly. At each time step, the model is evaluated in the parameter space with the distance to the real system parameters at the current time.
\vskip -1in
\paragraph{Results}
The trajectories and the error curves are presented in~Figure~\ref{figure:adaptive}.
Although the episodic agent initially follows the star, it eventually loses track of the dynamics because of the delay induced by planning over a time interval. Our adaptive policy, on the other hand, successfully explores the dynamics. 
Even though the planning of inputs has linear time complexity in~$T$ in both cases, we observe a slowdown by a factor 100 for the episodic agent.
We believe that this experiment on this toy model illustrates the relevance of an adaptive exploration policy in realistic settings.

\subsection{From exploration to exploitation}

The goal of exploration is ultimately to obtain an accurate model for the system for model-based control. In order to validate the relevance of our approach to this framework, we evaluate the model learned during exploration on model-based control tasks. We compare it to the recent active exploration algorithms RHC and MAX~\cite{pmlr-v100-schultheis20a,pmlr-v97-shyam19a} and to the model-free reinforcement learning algorithm SAC~\cite{pmlr-v80-haarnoja18b}.
\paragraph{Experimental setup} We experiment on the pendulum and the cartpole of the DeepMind control suite~\cite{tassa2018deepmind}, for which we added noise.
Throughout exploration, the learned model is evaluated using both ~\eqref{eq:performance_evaluation} and with the exploitation cost achieved by a model-based control algorithm on the swingup task.
The experimental details can be found in~Appendix~\ref{appendix:exploitaiton} and in~\cite{pmlr-v100-schultheis20a} along with the performance of the algorithms used for comparison.
The control task is considered solved when the cost is lower than a value that we have chosen arbitrarily based on simulations.
\paragraph{Results}
Our results are presented in~Table~\ref{table:exploitation} and in~Figure~\ref{figure:exploitation}. Our algorithm is sample-efficient in terms of exploitation, as it allows for a model-based control algorithm to solve the task faster than the other baselines. Its computational cost is low. 
\begin{table}[t]
    \vspace{-.3cm}
    \centering
    \caption{Number of observations required to solve the swingup task (first row) and average computation time per observation (second row) for the pendulum~(top rows) and the cartpole~(bottom rows).
    }
    \vskip 0.1in
    \begin{tabular}{l|lllll}
        {Method} & RAND & MAX    & SAC    & RHC   & FLEX
        \\
        \hline \hline
        samples  & {$>2$k} & 2000   & {$>2$k} & 500   & 50
        \\
        \hline
        compute  & 1      & 100    & 2      & 8     & 4
        \\
        \hline \hline
        samples  & {$>2$k} & {$>2$k} & {$>2$k} & $600$ & $300$
        \\
        \hline
        compute  & 1      & 20     & 1.5    & 2     & 1.6
        \\
        \hline
    \end{tabular}
    \label{table:exploitation}
    \vskip -0.15in
\end{table}

\section{Related work}

The pure exploration task has recently attracted much interest in both the control and the reinforcement learning communities.
Several methods have been proposed to learn the dynamics and to model uncertainty, including Gaussian processes~\cite{buisson2020actively}, Random Fourier Features~\cite{pmlr-v100-schultheis20a}, and neural networks ensembles~\cite{pmlr-v97-shyam19a, sekar2020planning}.
An online exploration policy for linear systems is studied in~\cite{greedy}. A theoretical study for the  identification of nonlinear systems can be found in~\cite{Mania2020}. The extension of optimal experiment design to neural networks is proposed in~\cite{mackay1992information} for static systems, and in~\cite{cohn1993neural} for dynamical systems with an offline algorithm and a focus on G-optimal designs.
\section{Conclusion}
We proposed an exploration algorithm based on D-optimal design, running online and adaptively. Our experiments demonstrate its sample efficiency both in terms of exploration and exploitation, its low computational cost and its ability to track time-varying dynamics. These results are encouraging for applications on real systems.

Although it is not the focus of our work, the online learning rule conditions the quality of exploration.
In particular, the agent should be able to learn with bounded memory to meet the computational requirements of embedded systems.
While we used a rather naive online learning algorithm, recent advances in the communities of machine learning and control are promising for learning dynamics online~\cite{9992939}. The computational cost of our method is dominated by the calculation of derivatives. Automatic differentiation is an active research field and we believe that progress in that direction can be made to reduce this cost.

It would be interesting to generalize FLEX to the more realistic setting of a partially observed state model~\cite{goodwin1977dynamic}. Another research direction is to use our exploration objective as an exploration bonus in the exploration-exploitation tradeoff.







\clearpage
\bibliography{references}
\bibliographystyle{icml2023}

\clearpage
\appendix
\onecolumn


\section{Key definitions and approximations}

We summarize step by step the approximations and the definitions pertaining FLEX from linear models to nonlinear models.  

\begin{table}[H]
    \centering
    \caption{Recap of our the important quantities we defined and their approximations for linear and nonlinear models.
    }
    \vskip 0.1in
    \begin{tabular}{l|cc}
         &Linear model& Nonlinear model
        \\
        \hline \hline
        Aassumption  & ${f}(z, \theta) = V(z) \times\theta + c(z)$ \quad as in \eqref{eq:linear-model}& ${f}(z, \theta)$ differentiable 
        \\
        \hline
        Learning  & online least squares     & online gradient descent 
        \\
          & $=$ maximum likelihood estimator     & $\simeq$ maximum likelihood estimator 
        \\
        & $\theta_{t+1}   = \theta_t - H_t \nabla \ell_t(\theta_t)$   \quad as in  \eqref{eq:ols_gradient}  & $\theta_{t+1}   = \theta_t - \eta_t \nabla \ell_t(\theta_t)$ \quad as in \eqref{eq:online-gd}
        \\
        \hline
         Feature map& $V(z)$ \quad  as in Definition \ref{definition:linear-model}  &
         $\displaystyle V(z) := \frac{\partial {f}}{\partial \theta}(z,\theta_\star)$ \quad as in \eqref{eq:jacobian_feature}
         \\
           & & $\theta_\star \simeq \theta_t$
        \\
          & $V_t := V(z_t)$   as in Definition \ref{definition:linear-model}   & $\displaystyle
         V_t := \frac{\partial {f}}{\partial \theta}(z_t, \theta_t)$ \quad as in \eqref{eq:regressor_nonlinear}
        \\
          & \multicolumn{2}{c}{$v := V^{(k)}$   as in Lemma \ref{lemma:determinant}}  
        \\
        \hline 
        Gram matrix  & \multicolumn{2}{c}{$M_t := \sum\limits_{s=0}^{t-1}\transp{V_s}V_s$ \quad as in \eqref{eq:gram}}
        \\
        \hline 
        Information matrix  & $I = M_t$  \quad as in \eqref{eq:fisher_exploration}& $I \simeq M_t$ \quad as in \ref{section:online-exploration_nonlinear}
        \\
        \hline 
        D-optimal information gain  & \multicolumn{2}{c}{$ G(z):=\log \det ( M_t + \transp{V(z)}V(z))$   \quad as in \eqref{eq:one-step_information}}
        \\
        \hline 
        Rank-one approximation  & \multicolumn{2}{c}{$  G(z) \simeq \log \det M_t +  \transp{v}(z) M_t^{-1} v(z)$   \quad as in \eqref{eq:D-optimal_rank-one}}
        \\
        \hline 
        Matrices  & \multicolumn{2}{c}{$ M:=M_t$},
        \\
          & \multicolumn{2}{c}{$D := \partial v / \partial x $}
        \\
          & \multicolumn{2}{c}{$B := \ud t \, \partial {f} / \partial u$ }
        \\
          & \multicolumn{2}{c}{$  Q := \transp{B} \transp{D} {M}^{-1} {D} B$}
        \\
        &\multicolumn{2}{c}{$b := -\transp{B} \transp{D}{M}^{-1}v$}
        \\
        &\multicolumn{2}{c}{as in \eqref{eq:regressor_nonlinear}}
    \end{tabular}
    \label{table:recap}
\end{table}


\section{ Exploration in high-dimensional environments}
\label{appendix:chain}

We propose an additional experiment showing the behaviour of our algorithm in high dimension. We will add this experiment to our submission.

The nonlinear system we consider is a chain of $N$ coupled damped pendulums, with unknown friction. Each pendulum is coupled with its two nearest neighbors. Only the first pendulum is actuated an the motion of the rest of the pendululms is due to the successive coupling~\cite{BITAR2017130}. Exploration of the system consists in finding the friction forces on the pendulums.

We believe that this system is illustrative for the typical setup of system identification. In robotics for example, the experiment seeks to measure the parameters of a humanoid constituted of large number of joints.
Furthermore, the system we propose poses both challenges of large dimension and underactuation. Indeed, the dimension of the state space is~$d(N) = 2N$, and only one of the~$N$ pendulums is actuated, the motion being propagated from neighbor to neighbor.
This experiment allows us to monitor both the sample efficiency and the computational cost of FLEX in a challenging setting of large dimension.

\paragraph{Setup}
The system is modeled with a linear model with unknown friction coefficients~$\theta \in \mathbb{R}^N$ from observations~$x_t \in \mathbb{R}^{2N}$.
We define the sample complexity as the number of samples required to obtain a~$10^{-2}$ parameter error. For different values of~$N$, we measure the sample complexity and the computational time of FLEX and Random.
\paragraph{Results}
We provide our resultsin~Table~\ref{table:chain}. These results capture the behaviour of the algorithm when the dimension~$d$ grows. Our algorithm FLEX seems to reach a linear sample complexity with respect to~$d$, with reasonable computational time, whereas random exploration fails to explore and yields super-linear sample complexity. Our results suggest that despite larger computational cost, FLEX remains sample efficient and competitive in high-dimensional environments.

\begin{table}[t]
    \centering
    \caption{Performance of algorithms FLEX and Random for the exploration of the~$N$ coupled pendulums.
    }
    \vskip 0.1in
    \begin{tabular}{l|lllll}
        $N$ & 2 & 5    & 10    & 20   & 50
        \\
        \hline 
        $d$  & 4  & 10  & 20  & 40  & 100 
        \\
        \hline \hline
        sample complexity, Random  & 20 & 100 & 200 & 500 & $>1000$ 
        \\
        \hline 
        compute, Random  & 1  & 2   & 6   & 60  & 250   
        \\
        \hline \hline
        sample complexity, FLEX & 10 & 50  & 100 & 200 & 500   
        \\
        \hline
        compute, FLEX  & 3  & 5   & 10  & 80  & 350   
    \end{tabular}
    \label{table:chain}
\end{table}


\section{Experimental details}
\label{appendix:details}
\subsection{Exploration benchmark}
\label{appendix:benchmark}
\subsubsection{Additional results}
\begin{figure}[H]
    \centering
    \includegraphics[width=.25\textwidth]{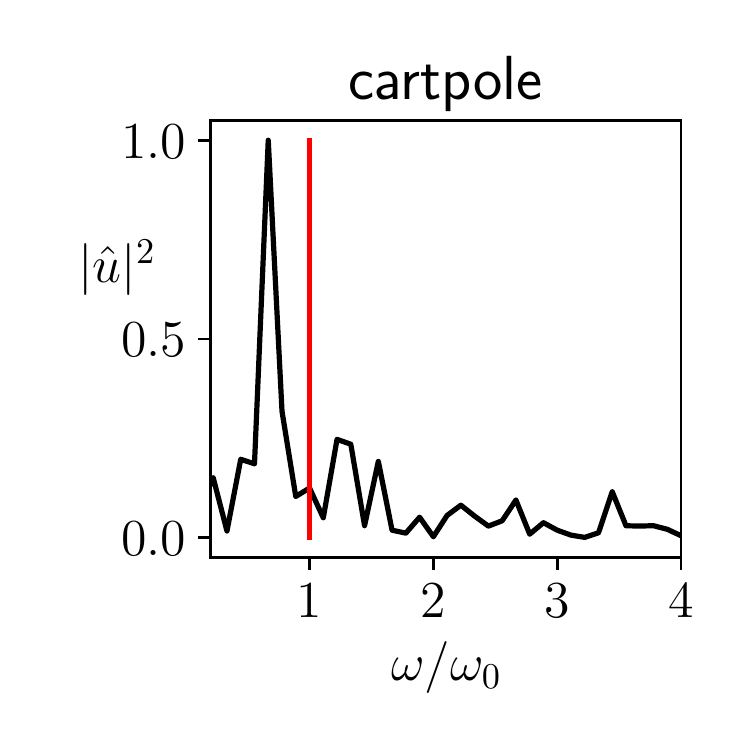}
        \caption{ Spectral density of the inputs generated by FLEX in the cartpole environment.}
    \label{figure:fourier}
\end{figure}

\todo{reference to additional results}
The temporal coherence of the inputs generated by FLEX is illustrated by~Figure~\ref{figure:fourier}. The spectral density shows that the energy is peaked on a frequency, hence implying that there is a temporal structure exciting the cartole near resonance, and thereby yielding informative trajectories.
\subsubsection{Environments}

We provide additional details on the environments and the learning models used. We denote the translation variables by~$q_x$ and~$q_y$ and the angle variables by~$q_\phi$.

\begin{sloppypar}
    \begin{environment}[Pendulum, $d=2, m=1$]
        \label{environment:pendulum}
        The dynamics are given by~\eqref{eq:pendulum-dynamics}.
    \end{environment}
\end{sloppypar}

\begin{environment}[Quadrotor, $d=6$, $m=2$]
    \label{environment:quadrotor}
    The planar quadrotor with nonlinear friction follows the following equations~\cite{zhang2014quadrotor}
    \begin{equation}
        \begin{aligned}
            m \ddot{q_x} & = - (u_1 + u_2) \sin q_\phi - \alpha q_x|\dot{q_x}| \dot{q_x}  \\
            m \ddot{q_y}  & =  (u_1 + u_2) \cos q_\phi -\alpha q_y  |\dot{q_y}| \dot{q_y} -mg \\
            I \ddot{q_\phi} & = r (u_1 - u_2)                                                 \\
        \end{aligned}
    \end{equation}
    with $m$ the mass and $I$ the moment of inertia.
\end{environment}

\begin{environment}[Cartpole, $d=4, m=1$]
    \label{environment:cart-pole}
   We implement the dynamics provided in~\cite{barto1983neuronlike}.
\end{environment}

\begin{environment}[Robot arm / double pendulum, $d=6$, $m=2$]
    \label{environment:arm}
    Equations available in \cite{chen2008chaos}.[$d=4$, $m=2$]
\end{environment}

\subsubsection{Baselines}
The random baselines returns $u_t \sim \frac{\gamma}{\sqrt{m}}\mathcal{U}([-1, 1]^m)$. The uniform policy maximizes the uniformity objective by gradient descent, with 100 gradient steps at each time step. The periodic policy returns inputs of the for~$u_t = \gamma \sin(\omega_0 t)$, with~$\omega_0$ an eigenmode of the system.
\subsubsection{Models}

We use the following learning models in~Section~\ref{section:benchmark}.  of width 8 with one hidden layer and $\mathrm{tanh}$ nonlinearity trained using ADAM optimizer~\cite{2015-kingma} with a batch size of 100.
\paragraph{Pendulum} We use the linear model of~Example~\ref{example:pendulum} and learn it by ordinary least squares.
\paragraph{Quadrotor} We learn the friction force with a neural net, and a learning rate of~$\eta=0.02$.
\paragraph{Cartpole} We parametrize~${f}(z, \theta) = a_\theta(\xi) + u\times b_\theta(\xi)$ with the observations~$\xi=(q_x, \dot{q}_x, \cos q_\phi, \sin q_\phi, \dot{q}_\phi)$, and~$a_\theta$ and~$b_\theta$ given by a neural network, trained with a learning rate of~$\eta=0.1$.
\paragraph{Arm} We use a neural network to learn~$f(., u=0)$ as a function of~$\xi=(\cos q_{\phi_1}, \sin q_{\phi_1}, \dot{q}_{\phi_1},\cos q_{\phi_2}, \sin q_{\phi_2}, \dot{q}_{\phi_2})$ and a learning rate of~$\eta=0.05$.
\subsection{Tracking of time-varying dynamics}
\label{appendix:time-varying}
The state of the spaceship in the plane is denoted
$x=   
        (q_x \, \dot{q}_x  \, q_y \, \dot{q}_y)\transp{}$. The center of the star has time-varying coordinates
        \begin{equation}
            \Big(\kappa_x(t), \kappa_y(t)\Big) = \Big(\cos(2\pi t/T), \sin(2\pi t/T)\Big).
        \end{equation}
        and the dynamics take the form 
\begin{equation}
    \label{eq:space_force}
\frac{\ud}{\ud t}
\begin{pmatrix}
    \dot{q}_x  \\ \dot{q}_y
\end{pmatrix}
=-\frac{1}{1+\frac{1}{\rho^2}\Big((q_x-\kappa_x)^2 + (q_y-\kappa_y)^2\Big)}
\frac{q}{\Vert q \vert_2}
\end{equation}
The agents know the dynamics down to the parameters~$\kappa_x, \kappa_y$ and~$\rho$, which they learn by online gradient descent, with optimier ADAM and a learning rate of~$\eta=0.01$.
\subsection{From exploration to exploitation}
\label{appendix:exploitaiton}

\begin{figure}[H]
    \centering
        \includegraphics[width=.25\textwidth]{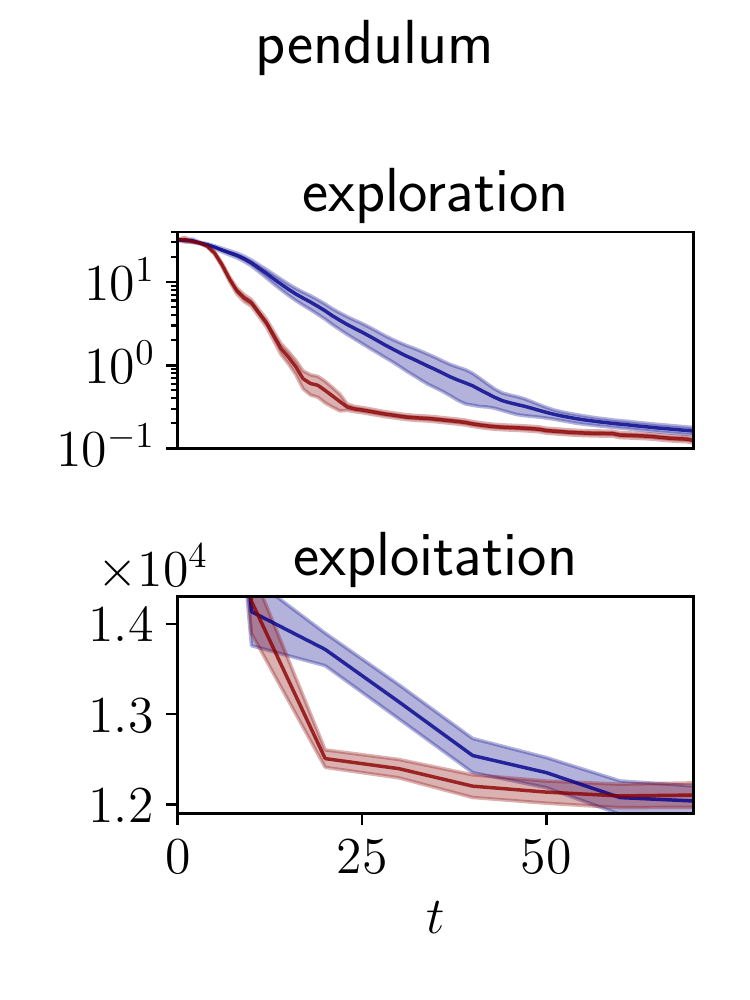}
    \includegraphics[width=.25\textwidth]{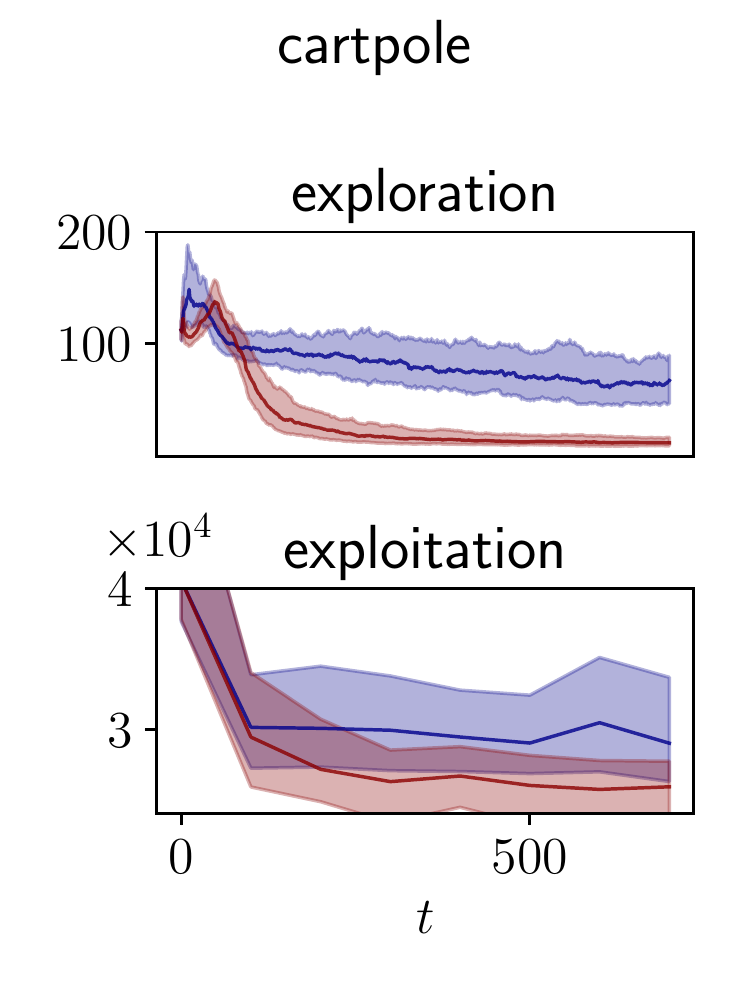}
    \raisebox{\height}{
        \includegraphics[height=1.5cm]{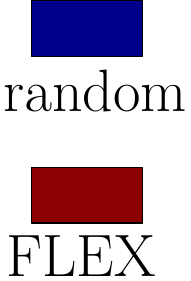}
        }
        \caption{Performance of our FLEX and random exploration evaluated on downstream model-based control tasks in pendulum and carptole.}
    \label{figure:exploitation}
\end{figure}

We add noise in the dynamics:~$\sigma=0.001$ for the pendulum and~$\sigma= 0.05$ for the cartpole.     

We use a linear model for the pendulum, and a neural network model with the same architecture as those of Section~\ref{appendix:benchmark} for the cartpole. The dynamics are learned as a function of $\cos q_\phi$ and $\sin q_\phi$, and $q_x$ and $\dot{q}_x$ for the cartpole.

Since we implemented our models with Pytorch~\cite{paszke2017automatic}, we used we use the \texttt{mpc} package and the iLQR algorithm for exploitation~\cite{amos2018differentiable}. The quadratic costs are 

\begin{equation}
    C=100(1-\cos q_\phi)^2 + 0.1 \sin^2 q_\phi + 0.1 \dot{q}^2_\phi + 0.001 u^2
\end{equation}
for the pendulum and 
\begin{equation}
    C=100 q_x^2 + 100(1-\cos q_\phi)^2 + 0.1 \sin^2 q_\phi^2 + 0.1 \dot{q}_x^2 + 0.1 \dot{q}_\phi + 0.001 u^2
\end{equation}
for the cartpole. When comparing the cost values to competitors, only the order  of magnitude matters since the control algorithm used for exploitation are different from an experiments to another.  

We measured the computational time on a laptop and averaged it over 100 runs for FLEX and the random policy, then compared with the values of~\cite{pmlr-v100-schultheis20a} and by setting the commputational time of the random policy to 1. Here again, only the orders of magnitude matter.  


\section{Proofs}
\subsection{Proof of Proposition~\ref{proposition:fisher_dynamics}}

\begin{proof}
    The data-generating distribution knowing the parameter~$\theta$ can be computed using the probability chain rule:
    \begin{equation}
        p(y|\theta) =( \frac{1}{\sqrt{2 \pi \sigma^2}})^t
        \exp \left(  - \frac{1}{2\sigma^2} \sum\limits_{t=0}^{t-1}
        \left\lVert  V(z_s) \times \theta - x_{s+1}\right\rVert^2_2 \right).
    \end{equation}
    which yields (omitting constants with respect to $\theta$)
    \begin{equation}
        \label{eq:log-likelihood}
        \begin{aligned}
            \ell(y, \theta)
             & = - \frac{1}{2\sigma^2} \sum\limits_{t=0}^{t-1}
            \left\lVert V(z_t) \times \theta - x_{t+1}   \right\rVert^2_2.
            \\
             & = - \frac{1}{2\sigma^2} \sum\limits_{j=1}^{d}\sum\limits_{t=0}^{t-1}
            \left\lVert v^{(j)}(z_t) \times \theta - x_{t+1}   \right\rVert^2_2.
        \end{aligned}
    \end{equation}
    Differentiating twice yields and taking the opposite yields
    \begin{equation}
        \begin{aligned}
            I(y, \theta) &=  \frac{1}{2\sigma^2} \sum\limits_{j=1}^{d}\sum\limits_{t=0}^{t-1}
             v^{(j)}(z_t) \transp{v^{(j)}(z_t)} 
            \\ 
            &=\frac{1}{2\sigma^2}\sum\limits_{t=0}^{t-1}\transp{V(z_s)}V(z_s)
            \\ 
            &=\frac{1}{2 \sigma^2} M_t.
        \end{aligned}
    \end{equation}
\end{proof}

\subsection{Proof of Proposition \ref{proposition:quadratic}}

\begin{proof}
    At first order in the neighborhood of $\bar{x} := x_t + \ud t \, {f}(x_t, 0)$,
    \begin{equation}
        \label{eq:expansion_g}
        v(x) = v(\bar{x}) + \frac{\partial v}{\partial x}(\bar{x}) \times  (x-\bar{x}) + \mathcal{O}\left(\Vert x - \bar{x}\Vert^2\right),
    \end{equation}
    Moreover, the first-order expansion of $x(u)$ near $u=0$ is
    \begin{equation}
        \label{eq:x_expansion}
        x(u) = \bar{x} + \ud t Bu + {o}\left(\ud t\right)
    \end{equation}
    where~$u$ is of order~$\gamma$. Substituting expansion~\eqref{eq:x_expansion} into expansion~\eqref{eq:expansion_g}  yields the first-order expansion of~$v(z(u))$. Omitting the additive constants, we obtain the stated result.
\end{proof}

\subsection{Proof of Proposition~\ref{proposition:numerical_solution}}

\begin{proof}
    See~\cite{greedy}. Since exploration aims at producing trajectories of large amplitude, we focus on inputs with maximal norm and assume an equality constraint in the optimization problem. The strict inequality can be handled readily with unconstrained minimization.
    Let us denote by~$u_\star$ a minimizer of~\eqref{eq:quadratic-problem},~$\{\alpha_i \}$ the eigenvalues of~$Q$, and~$u^{(i)}_\star$ and~$b_i$ the coordinates of~$u_\star$ and~$b$ in a corresponding orthonormal basis.
    By the Lagrange multiplier theorem, there exists a nonzero scalar~$\mu$ such that~$Qu_\star - b = -\mu \, u_\star$, where~$\mu$ can be scaled such that~$Q+\mu I_m$ is nonsingular. The inversion of the optimal condition and the expansion of the equality constraint yield
    \begin{subequations}
        \label{eq:optimality_conditions}
        \begin{align}
            u^{(i)}_\star = b_i/(\alpha_i + \mu)
            \\
            \sum_i \frac{{b_i}^2}{(\alpha_i + \mu)^2} = \gamma^2. \label{eq:norm}
        \end{align}
    \end{subequations}
    The minimizer~$u_\star$ is determined by solving~\eqref{eq:norm} for~$\mu$.
\end{proof}

\end{document}